\documentclass[journal]{IEEEtran}

\ifCLASSINFOpdf
\usepackage[pdftex]{graphicx}
\else
\fi

\IEEEoverridecommandlockouts
\usepackage{cite}
\usepackage{amsmath,amssymb,amsfonts}
\usepackage{algpseudocode}
\usepackage{algorithm}
\usepackage{graphicx}
\usepackage{graphicx,color}
\usepackage{textcomp}
\usepackage{amssymb}
\usepackage[utf8]{inputenc}
\usepackage{float}
\usepackage{amsmath}
\usepackage{hyperref}
\usepackage{tikz}
\usetikzlibrary{arrows,shapes,automata,petri,positioning,calc}

\tikzset{
    place/.style={
        circle,
        thick,
        draw=white,
        fill=white,
        minimum size=6mm,
    },
        state/.style={
        circle,
        thick,
        draw=white,
        fill=blue!20,
        minimum size=6mm,
    },
}

\def\BibTeX{{\rm B\kern-.05em{\sc i\kern-.025em b}\kern-.08em
    T\kern-.1667em\lower.7ex\hbox{E}\kern-.125emX}}
\begin{document}
\title{Decision-making and Fuzzy Temporal Logic\\
	\thanks{J. C. Nascimento is in  Department of Electrical  Engineering, Universidade Federal do Ceará, Sobral, Brazil.  e-mail:claudio@sobral.ufc.br}
	\thanks{This study was financed in part by the Coordenação de
Aperfeiçoamento de Pessoal de Nível Superior - Brasil (CAPES) -
Finance Code 001}
}
\author{José Cláudio do Nascimento}

\maketitle 

\begin{abstract}
This paper shows that the fuzzy temporal logic can model figures of thought  to describe decision-making behaviors. In order to exemplify, some economic behaviors observed experimentally were modeled from problems of choice containing time, uncertainty and fuzziness. Related to time preference, it is noted that the subadditive discounting  is mandatory in positive rewards situations and, consequently, results in the magnitude effect and time effect, where the last has a stronger discounting for earlier delay periods (as in, one hour, one day), but a weaker discounting for longer delay periods (for instance, six months, one year, ten years). In addition, it is possible to explain the  preference reversal (change of preference when two rewards proposed on different dates are shifted in the time). Related to the Prospect Theory, it is shown that the risk seeking and the risk aversion  are magnitude dependents, where the  risk seeking may disappear when the values to be lost are very high.
\end{abstract}


\section{Introduction}
\IEEEPARstart{F}{uzzy} logic addresses reasoning about vague perceptions of true or false \cite{zadeh1965fuzzy,zadeh1968probability,zadeh1988fuzzy}. Moreover, it represents very well our linguistic description about everyday dynamic processes  \cite{zadeh1976fuzzy, zadeh1975conceptI, zadeh1975conceptII, zadeh1975conceptIII, zadeh1996fuzzy,zadeh1983computational, efstathiou1979multiattribute}. Thus, the introduction of tense operators becomes a  necessity \cite{chajda2015tense, thiele1993fuzzy, moon2004fuzzy, cardenas2006sound, mukherjee2013fuzzy} because they are useful to represent the temporal indeterminism contained in some propositions \cite{prior2003time,macfarlane2003future}. For example, the hypothesis ``The weather is very hot'' has its  sense of truth revealed when we say ``The weather will always be very hot''. The  quantifier ``always'' communicates the belief about hypothesis confirmation in the future.

The task of developing a general theory for decision-making in a fuzzy environment was started  in  \cite{bellman1970decision}, but a temporal approach still needs to be discussed.
For example, do you prefer to receive  7 dollars today or to receive it 5 years later?
If you prefer today,   then you made a  distinction of the same value at distant points  from time, even though you have used vague reasoning about value changes  because we do not access  our equity value in memory to perform calculations. 

In decision-making, we do not only evaluate fuzzy goals, but also the time. The change caused by 7 dollars is strongly distinguishable between a  short time and a distant future. This suggests the existence of temporal arguments in fuzzy logic, where ``little change in wealth today'' is different from ``little change in wealth 5 years later''. The linguistic value ``little''  is irrelevant in this problem, but the arguments ``today'' and ``5 years later'' are decisive in the judgment.

The proposal here is to connect different fuzzy sets through time attenuators and  intensifiers. Hence, it is possible to simulate two figures of thought in hypotheses of  dynamic systems: meiosis to argue underestimated  changes and hyperbole to argue overestimated changes.

Through meiosis it is possible to formulate a concave expected change curve because it argues  minor changes for maximizing the  sense of truth. By this figure of thought within the fuzzy temporal logic it is noticeable that the hyperbolic discounting \cite{benzion1989discount, chapman1996temporal, chapman1995valuing, pender1996discount, redelmeier1993time, frederick2002time} and its subadditivity \cite{read2001time, read2003subadditive}, despite its numerous expected utility theory  anomalies \cite{loewenstein1992anomalies,ainslie2016cardinal,loewenstein1989anomalies}, is an intuitive dynamic  evaluation of wealth, where ergodicity is relevant in the problem \cite{peters2016evaluating, peters2011time}. 
Similarly, concave expected  change curve in lotteries \cite{bernoulli2011exposition} make certain outcomes more attractive than lotteries with uncertain outcomes.

Hyperbole  has an inverse reasoning to meiosis and it produces a convex expected change curve. Similarly, risky losses in the Prospect Theory have the same characteristic in its subjective value function \cite{tversky1986rational,tversky1992advances, kahneman2013prospect}. Then, it is shown here that the  risk seeking, which is a preference for risky losses rather than certain losses \cite{kahneman2013choices}, can be described in fuzzy environment by an imprecise perception for small losses. 
Thus, the indistinguishability between small negative changes leads to preference for hopes when only these are perfectly distinguishable.
On the other hand, when the losses are high, the risk seeking disappears and a kind of ruin aversion prevails \cite{taleb2018skin}, where it is better to lose a lot and stay with little than risk losing all means of survival after the lottery. 
In addition, the loss aversion behavior, where people prefer not to lose a given amount than to win the same amount \cite{kahneman2013prospect},  is interpreted by a disjunction between gains and losses hypotheses leading to the conclusion that such behavior is also amount dependent.

In essence, all the behaviors analyzed here are speculation examples   in dynamic systems, where we evaluate hypotheses and commit to outcomes before they emerge.
This paper shows, by modeling the Time Preference and the  Prospect Theory, that the fuzzy temporal logic allows to construct a rhetoric  for intertemporal and  probabilistic choices in fuzzy environment. The first problem takes us to focus on values and time and  the second focuses on values and probabilities. However, if the future is uncertain, there is no reason for time and uncertainty are studied in different matters \cite{lu2010many}. 
In addition, the feelings about judgments are amount dependents where the fuzziness can be decisive in some situations.  Therefore, time, uncertainty and fuzziness are concepts that can be properly studied by the fuzzy temporal logic in order to elaborate the decision-making rhetoric.

\section{Theoretical model}
This section provides a theoretical framework for the reader to understand as the figures of thought, meiosis and hyperbole, can be elaborated in the fuzzy temporal logic. In short, it is discussed the need of many-valued temporal logic, the existence of different temporal prepositions with  similar goals over time and, finally, it is shown as to perform the rhetorical development to make a judgment between two different hypotheses about the future. 

\subsection{Temporal and many-valued logic}
\label{TIL}
 Usually we  make decisions about dynamic processes whose states are unknown in the future. An amusing example can be found in an Aesop's fable, where the Grasshopper and the Ant have different outlooks for an intertemporal decision dilemma \cite{perry1965babrius}. In short, while the Ant is helping to lay up  food for the winter, the Grasshopper is enjoying the summer without worrying about the food stock. 
 
The narrative teaches about hard work, collaboration, and planning by presenting temporal propositions. These propositions have invariant meaning in time, but  sometimes they are true and sometimes false,  yet  never simultaneously true and false \cite{ohrstrom2007temporal}.  This property can be noted in the statement:
\[
\begin{array}{l}
D_1 =  \text{``We have got plenty of food at present'' \cite{perry1965babrius}.} 
\end{array}
\]
Although  $D_1$  has constant meaning over  time, its logical value is not constant. According to the fable, this statement is true in the summer, but it is false in the winter, hence there is a need to stock food.

If the logical value varies in time according to random circumstances, such as the weather of the seasons, how can we make inferences about the future? There seems to be a natural uncertainty that sets a haze over the vision of what will happen. For instance, consider the following statements about the Grasshopper: 
\[\centering
\begin{array}{rl}
D_2 = \text{``The Grasshopper stored a lot of food }\\ 
   \text{during the summer'';}\\
   D_3 = \text{``The Grasshopper stores food for winter''.}
\end{array}
\]   
According to the fable, we know that the statement  $D_2$ is false, 
but at the fable end, when the winter came, 
the Grasshopper says ``It is best to prepare for days of need''  \cite{perry1965babrius}. In this way, the truth  value in $D_3$ is ambiguous. We can not say with certainty that it is false, but 
we do not know how true it can be. Thus, we can only propose hypotheses to speculate the Grasshopper's behavior.

 A hypothesis is a proposition (or a group of propositions) provisionally anticipated as an explanation of facts, behaviors, or natural phenomena that must be later verified by deduction or experience. They should always be spoken or written in the present tense because they are referring to the research being conducted. In the scientific method, this is done independently whether they are true or false and, depending on rigor, no logical value is attributed to them. However, in everyday  language this rigor does not exist.  Hypotheses are guesses that argue for the decision-making before the facts are verified, so they have some  belief degree about the logical value that is a contingency sense performing a sound practical judgment concerning future events, actions or whatever is at stake.

Since there is no rigor in everyday speculations, then different propositions may arise about the same fact. If they are analyzed by binary logic, then we may have unnecessary redundancy that can lead to contradictions.
 However, the redundancy of propositions is not a problem within the speculation process, what suggests a many-valued temporal logic.
 
 We can discuss the limitations of a binary temporal logic by speculating the Grasshopper's behavior. For example, is it possible to guarantee  that the two hypotheses below are completely true simultaneously?
\[
\begin{array}{l}
\Theta = \text{``The Grasshopper stores a lot of food'';}\\
\theta = \text{``The Grasshopper stores little food''.}
\end{array}
\]
The hypotheses $\Theta$ and $\theta$  propose changes to different states. If $S_0$ is the initial state for stored food, the new state after storing a lot of food $M$ is $S_{\Theta}=S_0+M$. Analogously, the next state after storing little food is 
$S_{\theta}=S_0+m$ for $m<M$. 
Representing by $\Theta(t)$ the affirmation that $\Theta$ is true at the moment $t$ and considering the same initial state for both, then by binary logic
\[\Theta(t) \neq \theta(t) \text{ because } S_\Theta > S_\theta. \]
Therefore, affirming both as true leads to a  contradiction because the same subject can not simultaneously produce two different results on the same object. However, in a  speculation of the Grasshopper's behavior, none of the propositions can be discarded.

Evaluating by fuzzy logic, the linguistic variable ``stored food'' has values ``a lot of'' and ``little'' in the propositions  $\Theta$   and $\theta$. According to Bellman and Zadeh's model \cite{bellman1970decision}, these linguistic values  are the fuzzy constraints for inputs  $M$ and $m$ about the food supply. Meanwhile, the target states  $S_\theta$ and $S_\Theta$ can be the fuzzy goals.

An alternative development, but similar,  can be done through changes, what is in accordance with   human psychophysical reality \cite{kahneman2013choices}. Thus, in  this paper, the  goals  are the factors $ S_{\Theta}/S_0= 1+X$ and $S_{\theta}/S_0= 1+x$, where $X=M/S_0$ and $x=m/S_0$ are  changes. For example, in \cite{mukherjee2017loss}  the respondents were asked how they would feel gaining (or losing) a certain amount. It was noted that the emotional effects gradually increased as the amounts had grown. Therefore, the intensity of gains and losses are easily ordered by our perception, $x<X$.

The non-fuzzy set $-1\leq x<X$ for changes can have the fuzzifier effect  to transform it into a fuzzy set. Here, the general rule for fuzzy  goals  is simply  ``the bigger the better''. This rule can be represented by the membership function $\mu : [-1,\infty ) \to [0,1)$, where: 
\begin{enumerate}
\item $\mu(-1)=0$  means losing all;
\item if $-1\leq x<X$, then $\mu(x)<\mu(X)$, i.e., $\mu(x)$ is always growing;
\item $\lim\limits_{x\to \infty} \mu (x)=1$ means that the best gain goal (or change)  is undefined.
\end{enumerate}

\begin{figure}[!ht]
	\centering
	\includegraphics[scale=0.65]{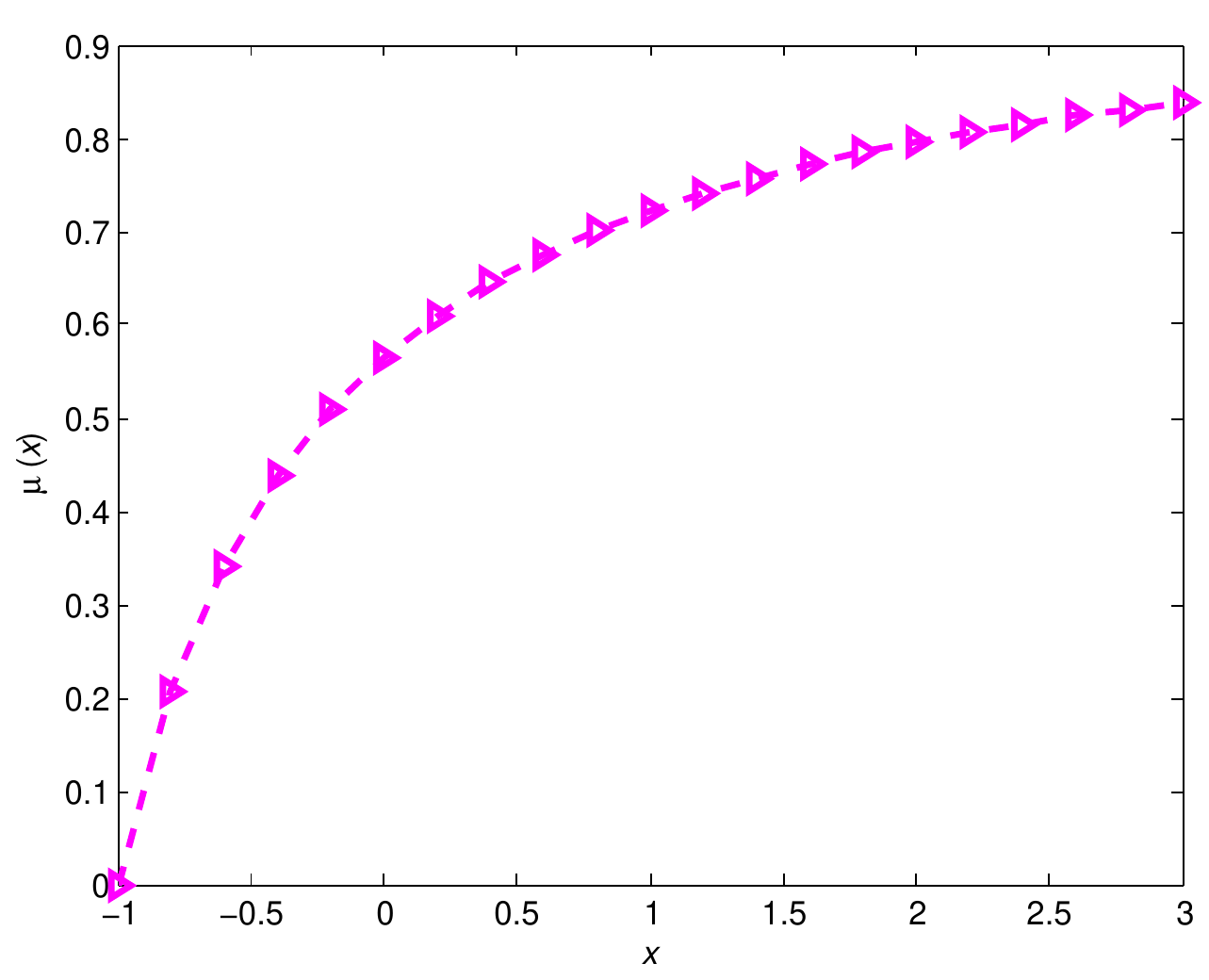}
		\caption{Membership function example $\mu(x)=1-[1-\alpha\beta(x+1)]^\frac{1}{\alpha}$ for $\alpha=-1.001$ and $\beta=1.3$. }	
	\label{figt}
\end{figure}

Figure \ref{figt} shows a membership function example that satisfies the above descriptions.   The membership function $\mu(x)$ is the general rule in decision-making because higher gains and lower losses, generally, cause better well-being. Therefore, others membership functions  will be a composition with the form $\mu(f(x))$, where $f$ must be in accordance with the rhetorical figure necessary to make a decision.

\subsection{Different hypotheses with similar future sentences} 
\label{HDSFS}
The  sense of truth that we have about a hypothesis becomes evident when we express it in the form of future statement. Given that a future sentence indicates an event occurrence after the moment of speech, then it can emphasize the fact realization with an intuitive implicit probability.
For example, consider the following future statements for
 $\Theta$ and $\theta$:
\[\centering
\begin{array}{rl}
F\Theta = \text{``The Grasshopper  will sometime store  }\\ 
   \text{a lot of food'';}\\
   G\theta = \text{``The Grasshopper  will always store}\\
     \text{ little food''.}
\end{array}
\]
The statement  $G\theta$ expresses  certainty with the modifier ``always'', while $F\Theta$ expresses uncertainty about the instant that the action occurs through the modifier ``sometime''. In temporal logic  $F\Theta$ is equivalent to saying which there is an instant  $t$ in the future where $\Theta$ is true, i.e, $\exists \;t$ such that $ (\text{now}<t) \wedge \Theta(t)$. 
Meanwhile,  $G\theta$ affirms  that $\theta$ is always true in the future,  $ \theta(t) \;\; \forall \: t>\text{now}$.

About the  certain sense, $F\Theta$ is known as the weak operator because  $\Theta$ is true only once in the future, while  $G\theta$ is known as the strong operator because it is true in all future periods.
If $\Theta$  does not always come true, then we can investigate its  sense of truth through the affirmative: 
\[
\begin{array}{rl}
GF\Theta = \text{``The Grasshopper  will frequently  store}\\
\text{ a lot of food''.}\\
\end{array}
\]
Where the quantifier ``frequently'' better argues for the  sense of truth of $\Theta$ because we have undefined repetitions in the same period in which $G\theta$ is true.

The frequency in which the propositions $\Theta$ and $\theta$ are true and the changes proposed by them determine the outcomes over time. 
The affirmative $GF\Theta$ communicates that the Grasshopper will frequently produce a strong change in the stored food stock  (change factor  $1+X$). On the other hand, $G\theta$ proposes a  small   change factor, $1+x$, but continuously over time. So what is the relation between $X$ and $x$  that generates a similarity of states between the two hypotheses over time? The relation that constructs this similarity can be obtained by the time average. Therefore, let us consider $\tau(t)$ as the total time where $GF\Theta$ is true and $t$ as the observation time. If $\Theta$ is true with a frequency given by
\begin{equation}
\lim\limits_{t\to \infty} \frac{\tau(t)}{t}=s,
\end{equation} 
then, the relation between  change factors  $1+x =(1+X)^s$, estimated by time average \cite{peters2016evaluating}, indicates that the sentences $GF\Theta$  and $G\theta$ have similar goals in the long run. This similarity is denoted in this work by

\begin{equation}
GF\Theta \sim G {\theta}.
\end{equation}

\begin{figure}[!ht]
	\centering
	\includegraphics[scale=0.8]{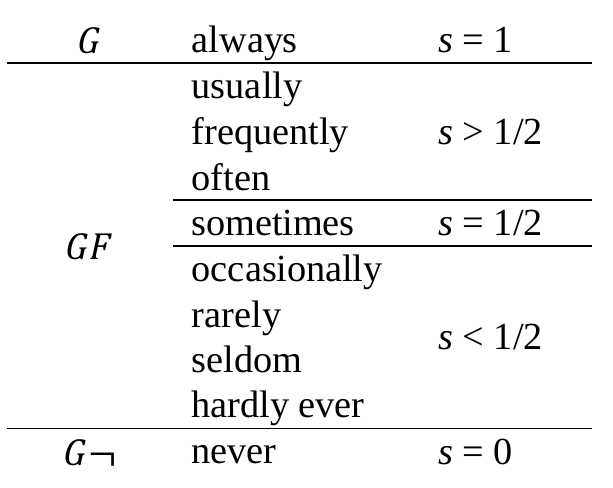}
		\caption{Adverbs of frequency   that can suggest the  sense of truth of $GF$.   The quantifier ``always'' indicates certainty,  while  ``never'' indicates impossibility.}	
	\label{fig1}
\end{figure}

The  sense of truth for the sentence $GF\Theta$ is quantified in the parameter $s$. It can be a stationary probability when
 $t$ is big enough, but we do not have much time to calculate this probability in practice.
In this way, we assume that the  sense of truth is an imprecise suggestion (intuition) for the time probability.
Figure \ref{fig1} presents some adverbs of frequency   that can suggest the  sense of truth  in future sentences.

The axiomatic system of temporal logic proposes that  
 $G {\theta} \Rightarrow GN {\theta}$, where  $N\theta$ stands for $\theta$ is true at the next instant. Therefore, the similarity  $GF\Theta \sim G {\theta}$ can be written by
\begin{equation}
  F\Theta \sim N {\theta},
\end{equation}
when $1+x \approx (1+X)^s$. 
Thus, the statement ``the Grasshopper will sometime store a lot of food'',  which have a  change factor $1+X$, is similar to the statement ``the Grasshopper will store little food at the next moment'', which has a  change factor $(1+X)^s$.

\subsection{Rhetoric: Meiosis and Hyperbole}
\label{MH}
In general, different hypotheses do not have similar changes in the future. For this reason, it is required that a rhetorical development  make a judgment between them. In this section, two figures  of thought are presented,  meiosis and hyperbole, which  can be used to compare hypotheses with different changes and  senses of truth. 

Still using Aesop's fable, imagine that we want to compare Ant and Grasshopper's performance with the following hypotheses:
\[
\begin{array}{l}
\phi = \text{``The Ant stores little food'';}\\
\Theta = \text{``The Grasshopper stores a lot of food''.}\\
\end{array}
\]
Assume that the Ant produces a  change   $y$ in its food stock while the Grasshopper produces a  change  $X$. If we think that the Ant is more diligent in its work than the Grasshopper, i.e., the  sense of truth for $\phi$ is maximum while the  sense of truth for $\Theta$ is ambiguous, so we can affirm $N\phi$  and $F\Theta$ in order to develop the following argumentation process:
\begin{enumerate}
\item elaborate a proposition $\theta$,  similar to $\Theta$,  which proposes a lower outcome, that has a  change  $x$ (for example, $\theta $ = ``The Grasshopper stores little food'');
\item express $\theta$ with maximum certainty in the future, $N\theta$ = ``The Grasshopper will store little food at the next moment'';
\item  calculate the relation $1+x \approx (1+X)^s$ to match the average changes between $N\theta$ and $F\Theta$ in order to obtain the similarity $F\Theta \sim N {\theta}$, where $X>x$ and $s\in [0,1]$;
\item finally, judge the affirmations $N\phi$ and $N\theta$ through fuzzy logic. In this specific problem we have
\[N\theta \text{ or } N\phi = \max\left\{\mu \left((1+X)^s-1\right),\mu(y)\right\}.\]
\end{enumerate}

\begin{figure}[!ht]
	\centering
	\begin{tikzpicture}[node distance=2cm and 1cm,>=stealth',auto, every place/.style={draw}]
         
    \node [state,initial text=,accepting by double] (1) [] {$\begin{array}{l}
 F\Theta \\
 \\
{\color{blue} F\Phi}
\end{array}$};  
    \node [state,initial text=,accepting by double] (0) [right=of 1] {$\begin{array}{l}
{\color{blue} N\theta} \\
\\
 N\phi
\end{array}$};
    \path[->] (1) edge [bend left] node {$\underbrace{(1+X)^s}_{\text{Meiosis}}$} (0);
    \path[->] (0) edge [bend left] node {$\overbrace{(1+y)^\frac{1}{s}}^{\text{Hyperbole}}$} (1);      
\end{tikzpicture}
		\caption{Diagram representing the meiosis and hyperbole procedure   for the judgment of hypotheses. The blue sentences $F\Phi$ and $N\theta$ are the  figures of thought.}	 
	\label{figMH}
\end{figure}
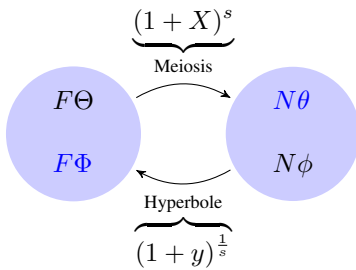

The above argument uses meiosis and the upper part of Figure \ref{figMH}  summarizes this procedure. 
In linguistic meiosis, the meaning of something is reduced to simultaneously increase something else in its place.
In the above mentioned case, proposing $\theta$ means reducing the stored  food change  by the Grasshopper.
At the same time, this suggests greater certainty because it makes the process more feasible in the future.
However, it is only a figure of thought to make an easy comparison because judging two sentences with the same  sense of truth, looking only at the change goals, is much simpler.

 The meiosis for Grasshopper's case  has a membership composite function given by
\[ \mu_{\text{ Grasshopper's goal}}= \mu \circ \mu_{\text{meiotic change}}(X)=\mu \left((1+X)^s-1\right).\] 
In general, $\mu_{\text{ Grasshopper's goal}}$ refers to the fuzzy goal ``the bigger the better is the change $(1+X)^s-1$ at the next moment''.  Like this, we can evaluate it for decision-making in real time.  

On the other hand, there is an inverse process to meiosis that is called hyperbole. Basically, it exaggerates a change in the outcome to reduce its  sense of truth. In everyday language, such statements are commonplace, as ``the bag weighed a ton''.
In this statement, we realize that  the bag is really heavy. However, is there a bag weighing really a ton? This is just a figure of speech.

In order to understand how to judge two hypotheses through a process of hyperbolic argumentation, consider again that we can make the future statements $F\Theta$ and $N\phi$ pass through the following steps:
\begin{enumerate}
\item elaborate a proposition $\Phi$,  similar to $\phi$,  which proposes a larger outcome, that has a change $Y$  (for instance, $\Phi $ = ``The Ant stores a lot of food'');
\item affirm $\Phi$ in the future with the same  sense of truth as the proposition $\Theta$, that is, $F\Phi$ = ``The Ant will sometime store a lot of food'';
\item  calculate the relation $(1+y)^{\frac{1}{s}}\approx 1+Y$ to match the average changes between $N\phi$ and $F\Phi$ in order to obtain the similarity $N\phi \sim F\Phi$, where $Y>y$ and $s\in [0,1]$; 
\item finally,  judge the fuzzy  changes goals of the affirmative $F\Phi$ and $F\Theta$. In this specific problem we have
\[F\Theta \text{ or } F\Phi = \max\left\{\mu(X), \mu\left((1+y)^\frac{1}{s}-1\right)\right\}.\]
\end{enumerate}

The hyperbole for Ant's case  has a membership composite function given by
\[ \mu_{\text{ Ant's goal}}= \mu \circ \mu_{\text{hyperbolic change}}(y)=\mu \left((1+y)^\frac{1}{s} -1\right).\] 
Thus, $\mu_{\text{ Ant's goal}}$ refers to the fuzzy goal ``the bigger the better is the change $ (1+y)^\frac{1}{s} -1 $ sometime in the future''.

 The bottom part of Figure \ref{figMH}  summarizes hyperbolic argumentation procedure. Note that the arguments by meiosis and hyperbole lead to the same conclusion. Therefore, they can be only two ways to solve the same problem. However, in section     \ref{PT}, where the Prospect Theory is evaluated, there may be preference for one of the methods according to the frame in which the hypotheses are inserted.

\section{Time Preference}
\label{PTemp}

Time preference is the valuation placed on receiving a good on a short date compared with receiving it on a farther date. A typical situation is  choosing to receive a monetary amount $m$ after a brief period (a day or an hour) or to receive $M>m$ in a distant time (after  some months or years).

The time   preference choice is a problem of logic about the future and in order to model it consider the following hypotheses:

\begin{itemize}
\item $\Theta_m =$ ``to receive $m$''  represents the receipt of the amount $m$ in short period, $t_m=t_0+\delta t$;
\item $\Theta_M =$ ``to receive $M$'' represents the receipt of the amount $M$ in longer time horizons, $t_M=t_0+\Delta t$. 
\end{itemize}
Each proposition has a  change factor for the individual's wealth. The proposition $\Theta_M$ has the change factor $(1+M/W_0)$, while $\Theta_m$ has the change factor $(1+m/W_0)$. Now, we perform the meiosis procedure for both hypotheses, reducing the changes and maximizing the  sense of truth:
\begin{itemize}
\item $N\theta_m =$ ``to receive an amount less than $m$ at the next moment''. This affirmative proposes a  change factor  $1+x_m$ in the individual's wealth;
\item $N\theta_M =$ ``to receive an amount less than $M$ at the next moment''. Similarly,  this affirmative proposes a  change factor  $1+x_M$.
\end{itemize}

The  senses of truth for the hypotheses $\Theta_M$  and $\Theta_m$ are  revealed when they are  affirmed in the future. Therefore, we have the following similarities:
\begin{itemize}
\item $F\Theta_M \sim N\theta_M$, if  $(1+x_M)\approx \left(1+\frac{M}{W_0}\right)^{s_M}$ ;
\item $F\Theta_m \sim N\theta_m$, if  $(1+x_m)\approx \left(1+\frac{m}{W_0}\right)^{s_m}$ .
\end{itemize}
Where $s_M$ and $s_m$ are the   senses of truth regarding the receipt of the values $M$ and $m$.
In this problem,  they cannot be time probabilities since the probabilistic investigation in this case is not convenient. However, intuitions about the realization of the hypotheses $\Theta_M$ and $\Theta_m$  are  feasible for individuals and they should be represented.

Now suppose, without loss of generality, that $M$ is large enough so that the individual prefers to receive it in the distant future. If we consider $n=\Delta t/\delta t$  periods in which $n$ attempts to receive $m$ until $M$'s receipt date are allowed, then we have
\begin{eqnarray}
\nonumber (1+x_M) &>& (1+x_m)^n \\
\Rightarrow\left(1+\frac{M}{W_0} \right)^{s_M} &>& \left(1+\frac{m}{W_0} \right)^{ns_m}.
\label{BhomBawerk}
\end{eqnarray}
Judging by fuzzy logic the two hyprothesis in the future, the ``or'' operation between change goals is indicated to finalize the meiosis procedure, 
\begin{eqnarray}
\nonumber \mu(x_M) &=& \max \left\{\mu(x_M), \mu \left( (1+x_m)^n-1 \right)\right\}\\
&=& \mu\left(\left(1+\frac{M}{W_0} \right)^{s_M}-1\right).
\end{eqnarray}

In general the time preference solution  is presented through a discount function. In order to use this strategy it is necessary to develop the same form on both sides of the inequality \ref{BhomBawerk}. For this, there is a value $\kappa$ such that $\kappa s_M > m$, where we can write
\begin{equation}
\left(1+\frac{M}{W_0} \right)^{s_M} = \left(1+\frac{\kappa {s_M}}{W_0} \right)^{ns_m}>\left(1+\frac{m}{W_0} \right)^{ns_m}.
\label{Passagem}
\end{equation}
The discount function undoes exactly the change caused by the proposition  $\Theta_M$, that is,
\begin{equation}
 \frac{1}{{\left(1+\frac{M}{W_0} \right)}} =\left(1+\frac{\kappa s_M}{W_0} \right)^{-\frac{s_m}{s_M}n} .
 \label{DescontoHiperbolico}
\end{equation}

\begin{figure}[!ht]
	\centering
	\includegraphics[scale=0.65]{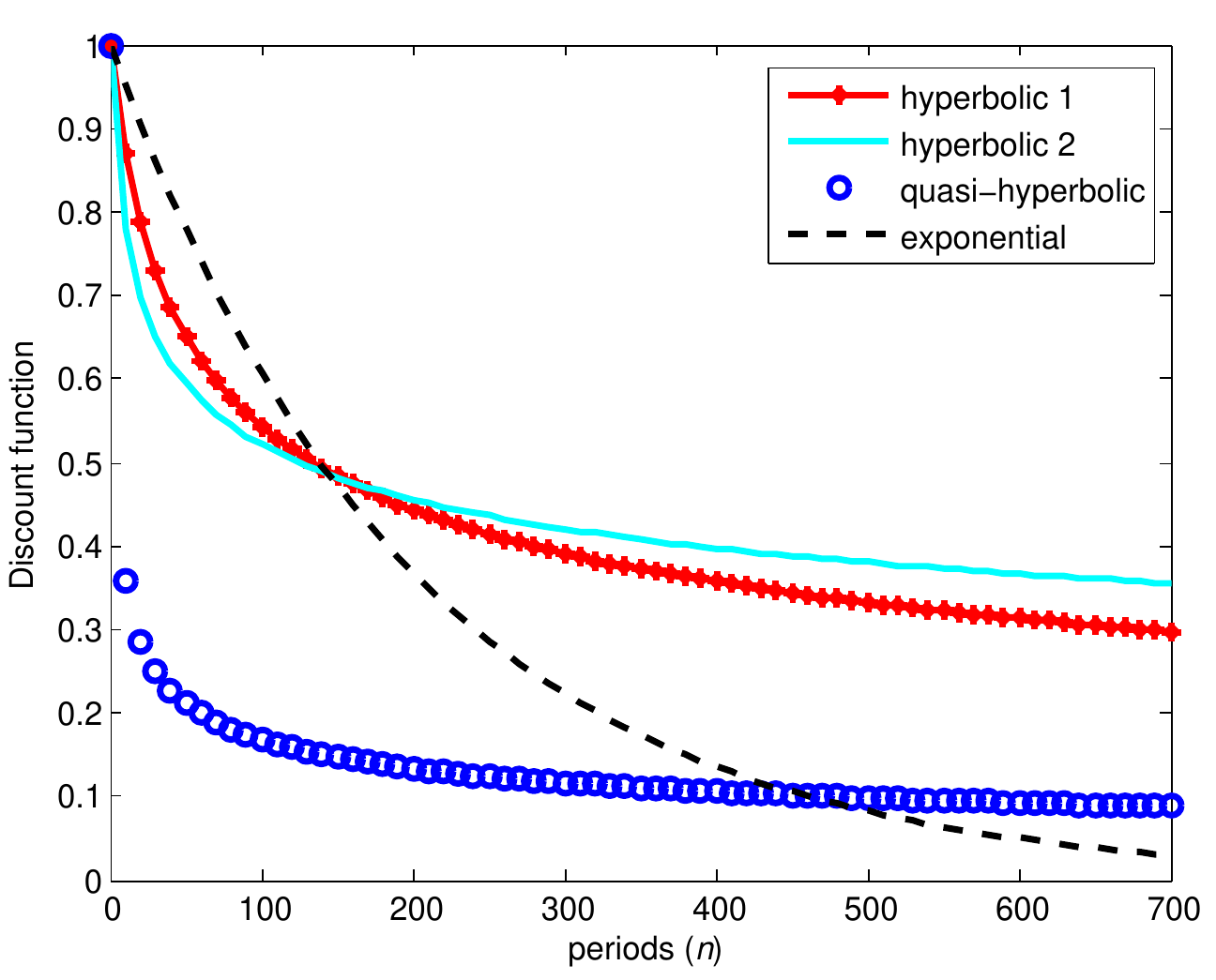}
		\caption{Discount function $e_h^{-\rho n}$ versus number of delayed periods: the dashed black curve is the exponential discounting for $\rho = 0.005$; the $\circ$-blue curve is quasi-hyperbolic discounting for $\rho = 0.7$ and $h = -3$; 
and the $\ast$-red and cyan curves are hyperbolic discounting for  $\rho_1 = 0.0175$ and $h_1 = -3$, and $\rho_2 = 0.05$ and $h_2 = -5$,  respectively.  }	
	\label{figDiscounting}
\end{figure}

Equation \ref{DescontoHiperbolico} describes the hyperbolic discount, the most well documented empirical observation of discounted utility \cite{frederick2002time}. When mathematical functions are explicitly fitted to experiment data, a hyperbolic shape fits the data better than the exponential form \cite{kirby1997bidding, kirby1995modeling, myerson1995discounting,rachlin1991subjective}. Among the functions proposed for the adjustment of experimental data, the discount function proposed by Benhabib, Bisin and Schotter \cite{benhabib2004hyperbolic}
\[e_h^{-\rho n}\equiv \left(1-h \rho n \right)^{\frac{1}{h}}\]
allows for greater flexibility of fit for exponential, hyperbolic, and quasi-hyperbolic discounting \cite{laibson1997golden} (see Figure \ref{figDiscounting}). In order to obtain it, we must reparametrize  equation \ref{DescontoHiperbolico} by doing 
\begin{eqnarray}
\frac{1}{h} &=& -\frac{s_m}{s_M}n,  \label{eqh} \\
\rho &=& \frac{\kappa s_m}{W_0}.
\end{eqnarray}

The parameter $h$ denotes  hyperbolicity and it gives the curve shape over time. For instance,  $e_h^{-\rho x}$ equals the exponential function $e^{-\rho x}$ when $h \to 0^-$. This means that there is plenty of time for possible trials with higher sense of truth until the date of the great reward. On the other hand,  $h\ll 0$ indicates time shortage for trials.  In theses cases, only the first periods have strong declines in the discount function. In short, equation \ref{eqh} shows that the senses of truth  and the time between rewards  determinate  the value of $h$.

 Furthermore, the discount rate $\rho$ quantifies the preference for goods and it is influenced by individual states of scarcity and abundance of goods. For instance, let us consider an individual called Bob. If an object is scarce for him (small $W_0$), then he places a higher preference (great $\rho$). Analogously, if $W_0$ represents an abundance state for him, he has a lower preference (small $\rho$).  This may cause great variability in experiments because the wealth distribution  follows the power law  \cite{levy1997new,druagulescu2001exponential,sinha2006evidence,klass2006forbes}. This means that $W_0$  can vary abruptly from one individual to another in the same intertemporal arbitrage experiment. 

\subsection{Discussion about time preference behaviors}

Daniel Read pointed out that a common evidence in time preference is the ``subadditve discounting'', in others words, the discounting over a delay is greater when it is divided into subintervals than when it is left undivided \cite{read2001time}. For example, in \cite{takahashi2006time,takahashi2008psychophysics} it has been argued that abstinent drug addicts may more readily relapse into addiction if an abstinence period is presented as a series of shorter periods, rather than a undivided long period. This property is present in the function $e_h^{-x}$, because if $h<0$, then
\[e_h^{-x} \; e_h^{-y}=e_h^{-x-y+h xy}<e_h^{-x-y} \text{ for all } x,y>0.\]

The hyperbolicity, due to intertemporal arbitrage, is always negative, $\frac{1}{h} = -\frac{s_m}{s_M}n$. Therefore, we have a subadditivity as a mandatory property in the time preference for positive awards.

However, the main experimentally observed behavior, hereinafter referred to as ``time effect'', concerns the observation that the discount rate  will vary inversely with the time period to be waited. Here, we can verify  this effect  as a consequence of the subadditivity found in the function $e_h^{-\rho n}$. After all, when we want to find the discount rate through a given discount $D$ after $n$ periods, then we calculate $-\frac{1}{n}\ln D $. For example, if we have an exponential discounting $e^{-\rho n}$, then $\rho= -\frac{1}{n}\ln e^{-\rho n} $. Thus, using the subadditive property we can develop 
\begin{eqnarray}
\nonumber \left(e_h^{-\rho}\right)^n\leq e_h^{-\rho n} & \Rightarrow & 
-\frac{1}{n}\ln \left(e_h^{-\rho}\right)^n\geq  -\frac{1}{n}\ln  e_h^{-\rho n} \\
\nonumber & \Rightarrow & \left\langle -\ln e_h^{-\rho} \right\rangle \geq   \left\langle -\frac{1}{n}\ln  e_h^{-\rho n} \right\rangle .
\end{eqnarray}
Therefore, if $h$ does not tend to zero from the left side, then the average discount rate over shorter time is higher than the  average discount rate over longer time horizons. 
For example, in \cite{thaler1981some} it was  asked to respondents how much money they would require to make waiting  one month, one year and ten years just as attractive as getting the \$ 250 now.  The median responses (US \$ 300,  US \$ 400 and US \$ 1000) had an average (annual) discount rate of 219\% over the one month, 120\% over the one year and 19\% over the ten years. Other experiments presented a similar pattern \cite{benzion1989discount, chapman1996temporal, chapman1995valuing, redelmeier1993time, pender1996discount}. Therefore, the time effect is a consequence of subadditivity when there is not plenty of time for trials at one of the hypotheses.

A second behavior,  referred to as  ``magnitude effect'', is also consequence of subadditivity. The reason for this is that magnitude effect and time effects are mathematically similar, because the discount rate is $\rho=\kappa s_m/W_0$ and  $ \kappa $ is growing for large values of    $ M $ (see equation \ref{Passagem}). In order to understand the similarity, note that the function $e_h^{-\rho n}$  varies with $ n \rho $. If we set the value $ n $, for example $ n = 1 $, and we vary the only rate $ \rho = r \rho_0 $, where $ \rho_0 $ is constant and $ r>1 $ is a multiplier which is growing for values of $ M $, then we have  the function $e_h^{-r\rho_0}$  analogous to $e_h^{-\rho n}$. Therefore, the magnitude effect made by $r$, similarly to the time effect,  results in
\[\left\langle -\ln e_h^{-\rho_0} \right\rangle \geq   \left\langle -\frac{1}{r}\ln  e_h^{-\rho_0 r} \right\rangle .\]

For example, in Thaler's investigation \cite{thaler1981some}, the respondents preferred, on average, \$ 30 in 3 months rather than \$ 15 now, \$ 300 in 3 months rather than \$ 250 now, and \$ 3500 in 3 months rather than \$ 3000 now, where the discount rates are 277\%, 73\% and 62\%, respectively. Other experiments have found similar results \cite{ainslie1983motives, benzion1989discount,green1994temporal,holcomb1992another,kirby1997bidding, kirby1995modeling, kirby1999heroin, loewenstein1987anticipation,raineri1993effect,shelley1993outcome, green1997rate}.

Another experimentally observed behavior is the ``preference reversal''. Initially, when individuals are asked to choose between one apple today and two apples tomorrow, then  they may be tempted to prefer only one apple today. However, when the same options are long delayed, for example, choosing between one apple in one year and two apples in one year plus one day, then to add one day to receive two apples becomes acceptable \cite{thaler1981some}. 

The preference reversal  shows how we evaluate value and time in the same hypothesis. For example, if $M_1<M_2$, then, in hypothesis, it is easier ``to receive  $M_1$ today'' than ``to receive $M_2$ today''. In this case, the sense of truth $s_1$ regarding the receipt of  $M_1$ is greater than the sense of truth $s_2$ regarding the receipt of $M_2$. For purposes of evaluation, consider that $ s_2 $ is not small enough to make $ M_1 $ preferable, that is
\begin{equation}
\left( 1+\frac{M_1}{W_0}\right)^{s_1} < \left( 1+\frac{M_2}{W_0}\right)^{s_2}.
\label{today}
\end{equation}

However, when one has to choose between the hypotheses ``to receive $M_1$ today'' and ``to receive $ M_2 $ tomorrow'', then it should be noted that sequential execution of these actions can give an advantage to receive $M_1$ in the future.
For simplicity, imagine that the proposal of each hypothesis happens  always  after receiving the reward. Then, in this case, the first hypothesis represents the attempt to receive $M_1$ every day, while the second hypothesis represents the attempt to receive $ M_2 $ every other day. If $M_1$ is not much less than $M_2$, then it will be more advantageous trying to receive $ M_1 $ twice  rather than expecting two days to receive $M_2$. Therefore, in dynamic situations, the waiting time between the hypotheses affects the frequency of attempts, resulting in
\begin{equation}
\left( 1+\frac{M_1}{W_0}\right)^{2s_1} > \left( 1+\frac{M_2}{W_0}\right)^{s_2}.
\label{Twice}
\end{equation}

On the other hand, when one has to choose between the hypotheses $ H_1=$``to receive $M_1$ in $n$ days'' and $ H_2=$ ``to receive  $M_2$ in $n+1$ days'', then a similar judgment can be made by evaluating the proposed action execution   over and over again over time. Since the waiting time to receive $ M_1 $ is shorter, then we can realize that the number of attempts to receive $M_1$ will be greater in the future ($(n+1)/n$ trials to receive $M_1$ for each trial to receive $M_2$). By fuzzy temporal logic, the choice between  $ H_1$ and $ H_2$ depends on the following result: 
\[ \max \left\{\left( 1+\frac{M_1}{W_0}\right)^{\frac{n+1}{n} s_1},\left( 1+\frac{M_2}{W_0}\right)^{s_2}\right\}.\]
When $n=1$, then it will be preferable to receive the reward $M_1$ (see equation  \ref{Twice}). This can also happen to other small values of $n$, for example, $n$ equals 2 or 3.  
However, when $n$ is large enough, the relation $(n+1)/n$ tends to  1 and  makes $M_2$ a   preferable reward (see equation \ref{today}). Thus, the preference between the rewards are reversed when they are shifted in time. In similar experiments, this behavior can be observed in humans \cite{kirby1995preference,green1994temporal,millar1984self,solnick1980experimental} and in pigeons \cite{ainslie1981preference,green1981preference}.

Hence, the time effect and magnitude effect on the discount rates, preference reversal and subadditivity are strong empirical evidences for the application of fuzzy temporal logic in intertemporal choices.

\section{Lotteries}
\label{PT}
In a  more realistic human behavior descriptive analysis with the psychophysics, the subjective value of lottery must be related to the wealth change \cite{kahneman2013choices}. However, changes after lotteries have incomplete information because we avoid calculating them using equity values, premiums and probabilities. Therefore, fuzzy sets are good candidates for representing these hypothetical changes. In addition, more realistic expectations should consider the evolution of outcomes over time \cite{peters2016evaluating}. Thus, the changes may have their expected values attenuated or intensified by the sense of truth (intuitive time probability).

\subsection{Meiosis and risk aversion}
 In order to model lotteries consider the hypothesis $\Theta_2$ = ``to win $M $'' and a probability $p \in [0,1]$. If $M>0$, then  by fuzzy temporal logic we can have:
\begin{itemize}
\item $l_1$ = to win $Mp$ (at the next moment);
\item $l_2$ = to win $M$ (at the next moment) with probability $p$.
\end{itemize}
The expression ``at the next moment'' does not appear in the experiments, but they are implicit because the low waiting time for the two lotteries seems to be the same. Moreover,  note that both lotteries are equivalents in the ensemble average ($E=pM$ for the two lotteries). 

Again, let us consider an individual called Bob who may repeat similar lotteries in the future. Therefore, this repetition can affect his decision \cite{peters2016evaluating}. If the lottery $l_2$ is repeated several times until he wins $M$, then $l_2$ is similar to affirm
\[F\Theta_2=\text{ ``will sometime win } M \text{'',}\]
where $p$ is the time probability (or sense of truth for lottery). Thus, if we perform the meiotic argumentation procedure  (see section \ref{MH}), then  the similar sentence which have equivalent outcomes to $F\Theta_2$ is
\[N\theta_2=\text{ ``will win } W_0\left(1+\frac{M}{W_0}\right)^p - W_0 \text{ at the next moment'',}\]
in which this sentence is a future affirmation of the hypothesis \[\theta_2= \text{ ``to win } W_0\left(1+\frac{M}{W_0}\right)^p - W_0 \text{''}.\]

Now, the  difference between $N\theta_2$ and $Nl_1$ consists only in the value of the award. Although values are reported in lotteries, variations on wealth are unknown because the cognitive effort to perform the division   $M/W_0$  is avoided. Thus, taking $x\geq 0$, such that   $x=M/W_0$, we can only evaluate  changes, 
\begin{eqnarray}
\nonumber N l_1  \text{ or } N\theta_2  &=& \max \{\mu(px),\mu\left((1+x)^p-1\right)\}\\
&=& \mu (px)  \text{ for all } x\geq 0.
\label{MeioseComp}
\end{eqnarray}

\begin{figure}[!ht]
	\centering
	\includegraphics[scale=0.65]{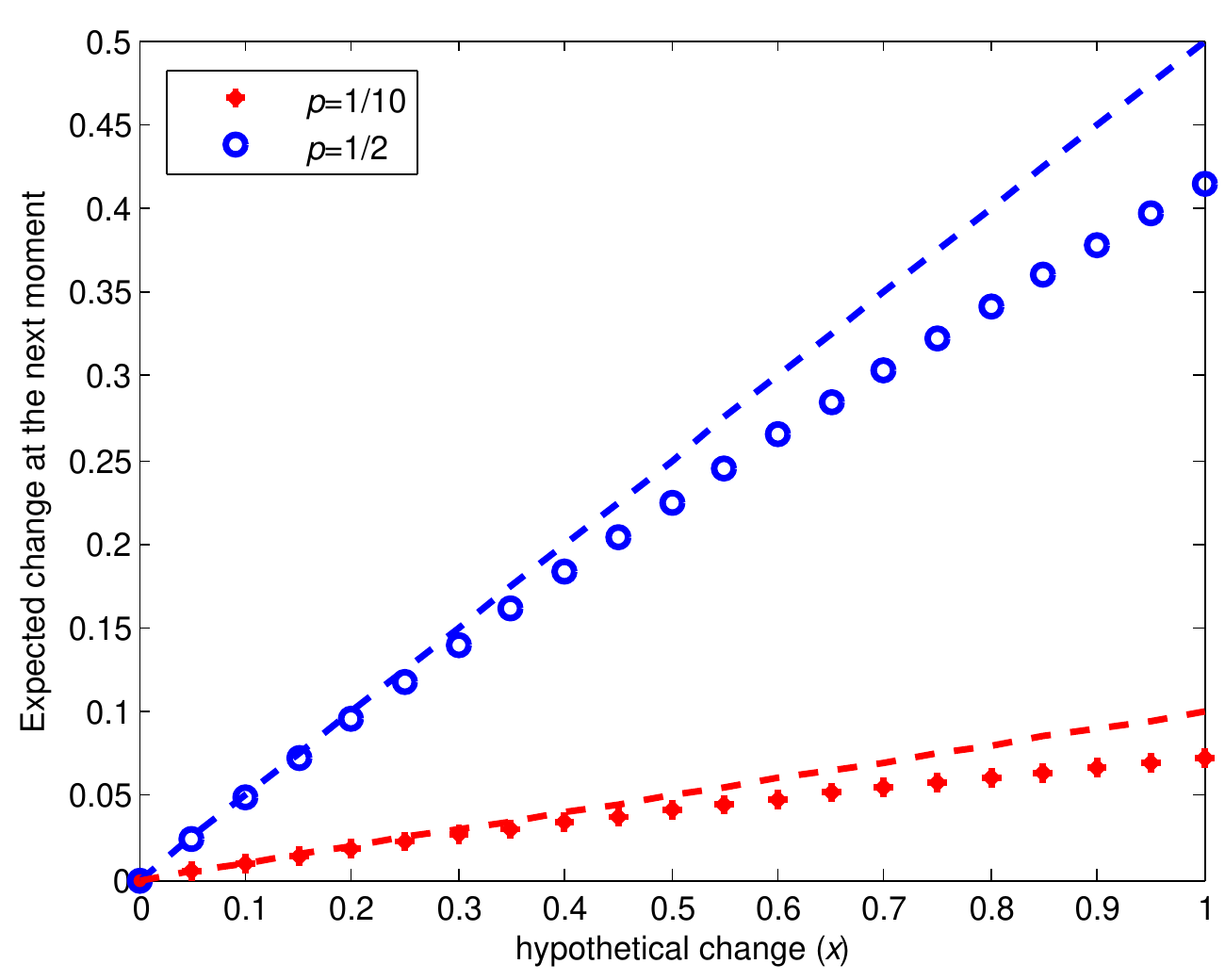}
		\caption{Function  ${\cal M}_p^+(x)$ to represent meiosis.   The blue dashed line  $x/2$ is tangent to the $\circ$-blue curve   given by ${\cal M}_{1/2}^+(x)$. Analogously, the red dashed line  $x/10$ is tangent to the   $\ast$-red curve given by ${\cal M}_{1/10}^+(x)$. Note that in the vicinity of zero the curves are close, so this is a region of low distinguishability for the changes.}	
	\label{figPT1}
\end{figure} 

The line $px$ is tangent to the concave curve $(1+x)^p-1$ at the point $x=0$ what results in  $px \geq (1+x)^p-1$ for $x\geq 0$.  An example can be seen in Figure \ref{figPT1} where the dashed blue line   $x/2$ is above the curve $\circ$-blue    $(1+x)^\frac{1}{2}-1$. Analogously, a similar illustration can be seen for $p=1/10$.  Thus, we can note that the lottery $l_1$ is preferable for any values of $M$ and $p$, because the line $px$  is always above the curve $(1+x)^p-1$. This result is consistent with the respondents' choices in the Kahneman and Tversky experiments  \cite{kahneman2013prospect}. Thus, the concave curve representing the expected positive change at the next moment can be described by
\begin{eqnarray}
\nonumber {\cal M}^+_p(x)&=& (1+x)^p -1  \\
&=& p\ \text{ln}_p (1+x) \text{ for all } x\geq 0.
\label{MeioseMais}
\end{eqnarray}
The function $\text{ln}_p (x)\equiv (x^p -1)/p$ is defined here as in \cite{nivanen2003generalized} and it is commonly used in nonextensive statistics \cite{tsallis1988possible,tsallis1999nonextensive}.

\subsection{Hyperbole and risk seeking}
 Kahneman and Tversky also show that the subjective value function  is not always concave \cite{kahneman2013choices}. They noted that in a loss scenario there is a convexity revealing a preference for uncertain loss rather than for certain loss.
   If we replace the word ``win'' for ``lose'' in the lotteries $l_1$ and $l_2$, then we have the following lotteries that result in the wealth decrease:  
\begin{itemize}
\item $l_3$ = to lose $Mp$ (at the next moment);
\item $l_4$ = to lose $M$ (at the next moment) with probability $p$.
\end{itemize}

Now consider the hypothesis $\Theta_4$ = ``to lose $M$''. If $l_4$ is repeated until the individual loses $M$, then this lottery becomes similar to 
\[F\Theta_4=\text{ ``will sometime lose } M \text{'',}\] 
where $p$ is the time probability   (sense of truth for $F\Theta_4$). 

Simultaneously, the lottery $l_3$ proposes a certain loss. Then we can reduce its sense of truth for $p$ to compare with $F\Theta_4$. For this, let us consider the following hyperbole
 \[L_3 = \text{``to lose } W_0-W_0\left(1-\frac{pM}{W_0}\right)^\frac{1}{p}\text{'',}\]
in which the affirmation in the future
\[\centering
\begin{array}{rl}
FL_3  = \text{``will sometime lose } W_0-W_0\left(1-\frac{pM}{W_0}\right)^\frac{1}{p}\text{''}
   \end{array}
\]
arguments an expected change  $(1+px)^\frac{1}{p}-1$ for $-1\leq x<0$.

\begin{figure}[!ht]
	\centering
	\includegraphics[scale=0.65]{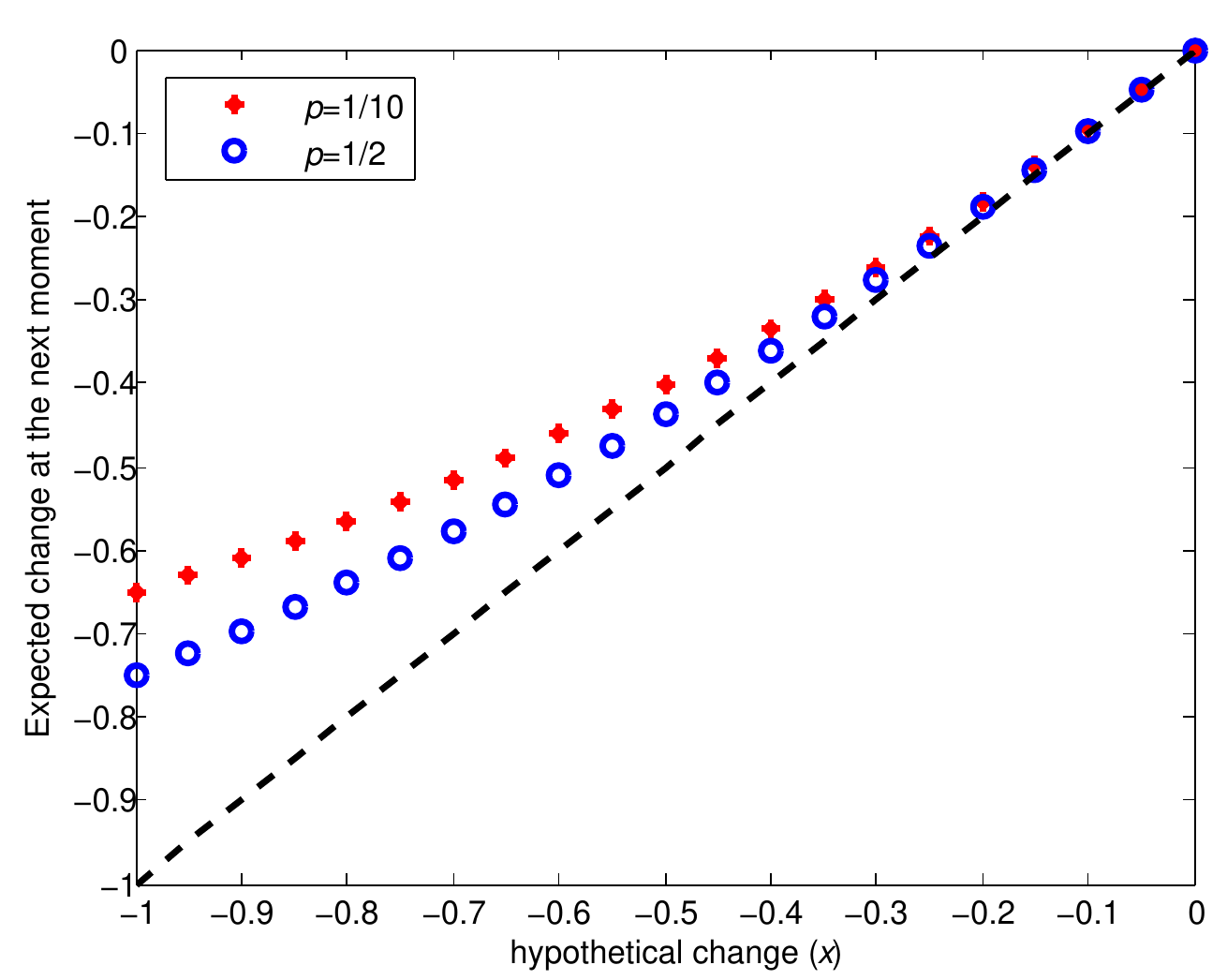}
		\caption{The hyperbolic curves $(1+px)^{\frac{1}{p}}$ for $p=1/2$ and $p=1/10$. Note that the curves tangentiate the black dashed line $x$ at the point zero. The interval $-0.1\leq x <0$  is a region of low distinguishability. }	
	\label{figPT2}
\end{figure}

 The line $x$ is tangent to the convex curve $(1+px)^\frac{1}{p}-1$ at the point $x=0$ for any  $p$, what results in  $(1+x)^\frac{1}{p}-1 \geq x$ for $-1\leq x<0$. In Figure \ref{figPT2}  the dashed black line   $x$  represents the proposed  change in $l_4$ and the curves  $\ast$-red and $\circ$-blue, 
belonging  to the  family of curves $(1+px)^\frac{1}{p}-1$,  represent the hyperbolic argumentation for $l_3$.
Therefore, note that  the family  of curves $(1+px)^\frac{1}{p}-1$   is very close to  line $x$ until 0.1. 
This means that they have low distinguishability in this region, in other words, uncertain and certain losses can be imperceptible changes  when the losses are small.
In fuzzy logic is equivalent  to choose between ``small decrease in wealth with certainty'' and ``small decrease in wealth with probability $p$''. 
The decreases in wealth are almost the same and undesirable, but the uncertainty argues hope for escaping losses and it is desirable. Thus, the uncertain option for losses will be more attractive in this situation. Then, in order to simulate risk seeking  in the losses region   we must insert a rate $\rho$ into the hyperbolic argumentation process, so that
\begin{eqnarray}
\nonumber {\cal H}^-_p(\rho x)&\equiv &(1+p\rho x)^\frac{1}{p}-1\\
\nonumber &=& e_p^{\rho x}-1.
\end{eqnarray}

\begin{figure}[!ht]
	\centering
	\includegraphics[scale=0.65]{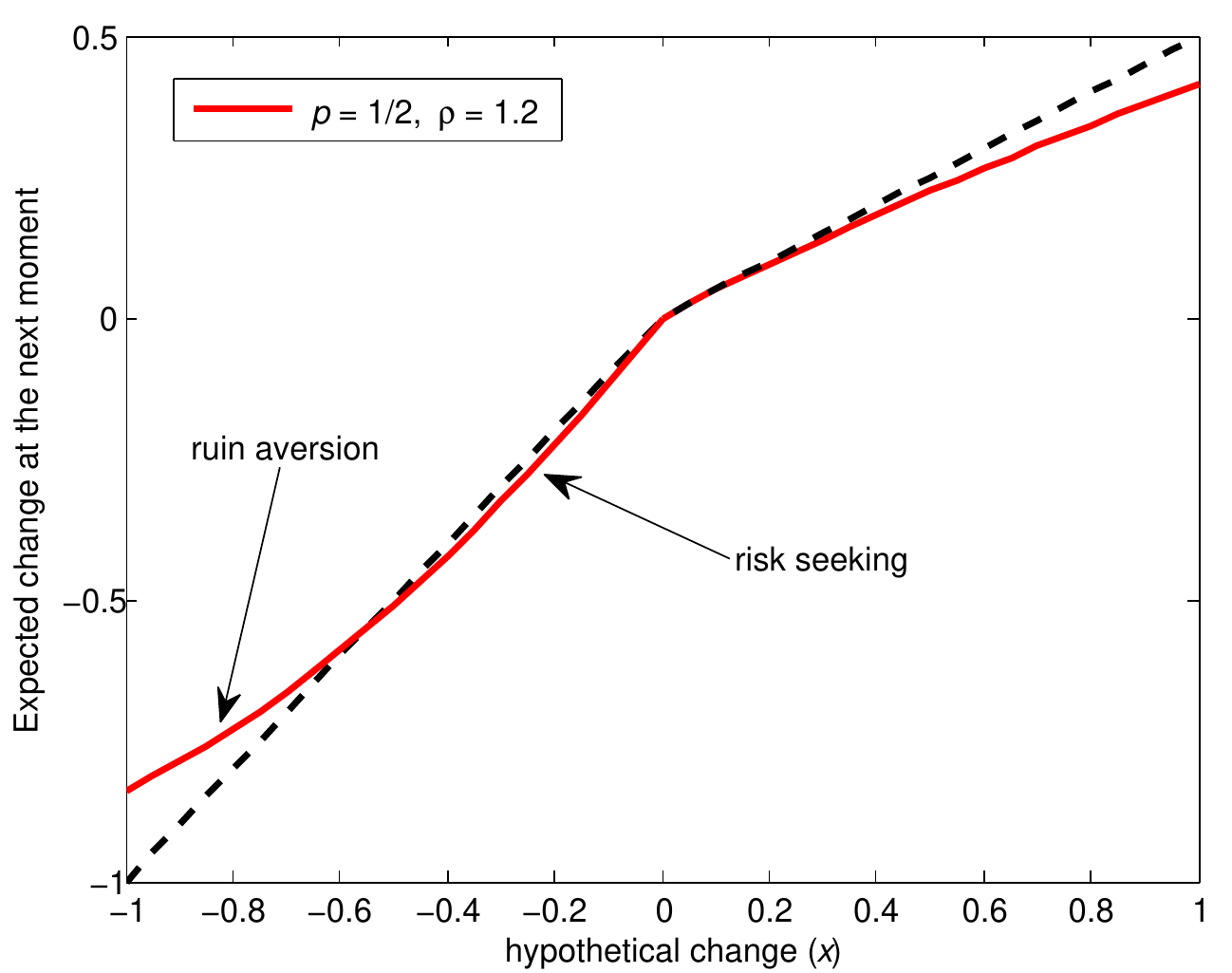}
		\caption{Function ${\cal S}_{0.5} (x)$ for $\rho=1.2$. When the red curve is below the dashed black line we have risk seeking (interval $-0.55<x<0$). On the other hand, we have a kind of ruin aversion when the red line is above the line $x$ (interval $-1<x<-0.55$). The risk aversion behavior is always present for all $x>0$ (dashed black line $x/2$ above the red curve). }	
	\label{figPT3}
\end{figure}

The rate $\rho$ makes the curve ${\cal H}^-_p(\rho x) $ more convex. Thus, its first values pass  below the line $x$ to simulate the risk seeking.  In Figure \ref{figPT3} the red curve   has  $\rho=1.2$ and $p=1/2$ to simulate this effect in the interval $-0.55<x<0$. The value $\rho=1.2$ was exaggerated just to make the risk seeking visible in the figure, but it should be slightly larger than 1 for a more realistic representation.

Now we can make a judgment between the sentences $FL_3$ and $F\Theta_4$ because the difference between them is only the argued decrease  wealth, 
\begin{eqnarray}
 FL_3 \text{ or } F\Theta_4 &=& \max\{ \mu \left( e_p^{\rho x}-1\right), \; \mu(x)\}  \label{CompHiper}\\
\nonumber &=& \left\{ \begin{array}{l}
 \mu(x)  \text{ for small losses},\\
\mu \left( e_p^{\rho x}-1\right) \text{ for big losses}.
\end{array}\right.
\end{eqnarray}
This means that the lottery $l_4$ is preferable when $M$ represents small losses, but when the losses are large, then $l_3$ is interpreted as the best option. Thus, the risk seeking disappears when the fuzzy sets are distinguishable.

 In order to understand the disappearance of  risk seeking, imagine all the goods necessary for your survival. After imagining them, then what do you prefer?  ``to lose 50\% of all assets'' or ``a chance to lose all assets with probability 0.5''? If you chose the first option then you understood that a ruin aversion \cite{taleb2018skin} can be dominant in this context. On this account, if   ``all assets'' is representing all the means of survival, then the second option may mean a death anticipation  after the lottery, while the former will allow the continuation of life with half of all assets.

Finally, the function with expected changes at the next moment, convex for losses and concave for gains, exhibiting a S-shape, can be defined by
\[{\cal{S}}_p(x)=\left\{ \begin{array}{ll}
 p\ \text{ln}_p (1+x) & \mbox{if}\quad x\geq 0,\\
e_p^{\rho x}-1& \mbox{if}\quad -1<x<0.
\end{array}\right.\]
In Figure \ref{figPT3}, when  $p=1/2$ and $\rho=1.2$, the function ${\cal{S}}_p(x)$  has risk seeking for $-0.55<x<0$, but has a kind of ruin aversion for $-1<x<-0.55$. The risk aversion is always present for all $x>0$.

\subsection{Loss aversion and disjunction between hypotheses}
\label{averisco}
The loss aversion principle, ${\cal{S}}_p(x)<-{\cal{S}}_p(-x)$, refers to the tendency to avoid losses rather than to acquire equivalent gains, in other words,  it is better not to lose \$ 1,000  than to win \$ 1,000.  

In order to understand why the curve is steeper on the losses side, consider that Bob has only \$10,000 (all of his money). If now he loses \$ 1,000, then the variation is -10\% and his new wealth is \$ 9,000. However, if he wins \$ 1,000,  then the positive variation is 10\%  and his new wealth is \$ 11,000. So far this process seems fair, but we need to look at it dynamically.
 If he will have \$ 9,000 at the next moment, then he will need to gain 11.11\% to get back \$ 10,000. On the other hand, if  he will have \$ 11,000 at the next moment, then the required variation  is -9.09\%  to get back \$ 10,000. So which is the most difficult change to happen? Exactly, the 11.11\% that restore the previous state after to lose 10\%. Therefore, it is better not lose 10\% than to gain 10\% in the long-term gamble.

This behavior can be modeled in fuzzy temporal logic through the disjunction operator. In order to understand the details about the disjunction between loss and gain hypotheses, consider the lottery    
\begin{eqnarray}
\nonumber L_{wl} &=& \text{``to win } M_1 \text{ with probability } p\\
\nonumber &\text{ }&\text{ or to  lose }M_2\text{ with probability } q\text{''.}
\label{AversoRisco}
\end{eqnarray}
If winning  $M_1$ produces a gain  $x_1>0$ and losing $M_2$ produces a loss, $-1\leq x_2 <0$, so we have the following atomic hypotheses
\[
\begin{array}{l}
H_1 = \text{``to win }M_1\text{'' and } H_2 =  \text{``to lose }M_2\text{''.}
\end{array}
\]
Where the sense of truth for $H_1$ is $p$ and the sense of truth for $H_2$ is $q$. The future statement for disjunction is 
\begin{eqnarray}
\nonumber F (H_1 \vee H_2) &=& \text{``to win } M_1 \text{ or to  lose }M_2\\\nonumber &\text{ }&\text{once in the future''.}
\label{LoteriaDisjunta}
\end{eqnarray}
The change average in this disjunction is $(1+x_1)^p(1+x_2)^q$  for $p+q<1$. This means that one of the hypotheses may be true at the next instant, or none, because only one will sometime be true  in the future.
 
 Another way of affirming a disjunction of losses and gains  is ensuring that one or the other will be true at the next instant, $N (H_1 \vee H_2)$. The change average in this case is  $(1+x_1)^p(1+x_2)^q$  for $p+q=1$. This means that $H_1$ or $H_2$ will be true at the next moment with absolute certainty. Uncertainty is just ``which hypothesis is true?''. Therefore, the judgment preceding the decision whether or not  participating in this lottery, $N(H_1 \vee H_2)$ or  nothing, is equal to
\begin{equation}
\nonumber  \max \{\mu\left((1+x_1)^p(1+x_2)^q-1\right),\mu(0)\}.
\end{equation}
The lottery $L_{wl}$, which is a  loss and gain disjunction, will be considered fair if the parameters $x_1$, $x_2$, $p$ and $q$  guarantee $(1+x_1)^p(1+x_2)^q-1>0$. In the experiment described at \cite{kahneman2013prospect}, the value $p=1/2$ and $x_1=x_2=x$ generate the inequality $\sqrt{1-x^2}-1<0$  that makes the lottery unfair. Thus, the respondent's choice for not betting seems to reveal a perception about the lottery dynamics. In addition, it can be noted  that  expected negative change has its intensity increased by $x$. Therefore,   the intensity of loss aversion is amount magnitude dependent. This means that the feeling of aversion to the lottery increases with the growth of the amount. In \cite{mukherjee2017loss} is presented an empirical evidence of this behavior.

\section{Conclusion}
Heuristics are cognitive processes that ignore part of the information and use simple rules to make a decision quicker and easier. In general, they are defined as strategies that seek alternatives, stop searches in a short time, and make a decision soon after.

Within heuristic processes, some decision-making requires the hypotheses judgment about dynamic processes before they take place in time. Time Preference Problem and Prospect Theory are famous examples. The first evaluates the goods receipt at different future dates  and the second requires lotteries valuation before their outcomes are known. The common characteristics  between these two problems noted here  are the magnitude dependence and the inseparability between time and uncertainty. On the magnitude dependence it can be concluded that:
\begin{itemize}
\item the magnitude effect in time preference is a consequence of subadditivity;
\item the risk seeking can disappear in the Prospect Theory if  high magnitude losses were considered. In addition, the aversion risk  increases with the growth of the amounts.
\end{itemize}
On the other hand, on the  inseparability between time and uncertainty it can be concluded that:
\begin{itemize}
\item in the time preference problem, the number of uncertain trials for the short-term hypotheses until verification of the long-term hypothesis produces the subadditive discounting, and consequently, higher annual average rates as the waiting time decreases. In addition, the preference reversal  occurs because the number of allowed trials is changed when the  hypothesis deadlines are shifted in time;
\item the probabilities of lotteries represent the temporal indeterminism about the future. Thus, the S-shaped curve in the Prospect Theory can be described by expected fuzzy changes of temporal hyptotheses.
\end{itemize}

If the future is uncertain, then time and uncertainty about changes can not mean two independent matters. For this reason, choice under uncertainty and intertemporal choice, traditionally treated as separate subjects, are unified in a same matter in this paper to elaborate the rhetoric for the decision-making. 

In addition, it is shown that the fuzziness can changes to prospective judgments about  magnitude dependent gains and losses. This means that  a given problem may have different decisions simply by changing  the values of the rewards, even if time and uncertainty context are not changed. 
Exactly in these situations, fuzzy environment modeling will be essential to represent the decision-making.

\bibliographystyle{IEEEtran}
\bibliography{ref}

\begin{thebibliography}{10}
\providecommand{\url}[1]{#1}
\csname url@samestyle\endcsname
\providecommand{\newblock}{\relax}
\providecommand{\bibinfo}[2]{#2}
\providecommand{\BIBentrySTDinterwordspacing}{\spaceskip=0pt\relax}
\providecommand{\BIBentryALTinterwordstretchfactor}{4}
\providecommand{\BIBentryALTinterwordspacing}{\spaceskip=\fontdimen2\font plus
\BIBentryALTinterwordstretchfactor\fontdimen3\font minus
  \fontdimen4\font\relax}
\providecommand{\BIBforeignlanguage}[2]{{%
\expandafter\ifx\csname l@#1\endcsname\relax
\typeout{** WARNING: IEEEtran.bst: No hyphenation pattern has been}%
\typeout{** loaded for the language `#1'. Using the pattern for}%
\typeout{** the default language instead.}%
\else
\language=\csname l@#1\endcsname
\fi
#2}}
\providecommand{\BIBdecl}{\relax}
\BIBdecl

\bibitem{zadeh1965fuzzy}
L.~A. Zadeh \emph{et~al.}, ``Fuzzy sets,'' \emph{Information and control},
  vol.~8, no.~3, pp. 338--353, 1965.

\bibitem{zadeh1968probability}
L.~A. Zadeh, ``Probability measures of fuzzy events,'' \emph{Journal of
  mathematical analysis and applications}, vol.~23, no.~2, pp. 421--427, 1968.

\bibitem{zadeh1988fuzzy}
------, ``Fuzzy logic,'' \emph{Computer}, vol.~21, no.~4, pp. 83--93, 1988.

\bibitem{zadeh1976fuzzy}
------, ``A fuzzy-algorithmic approach to the definition of complex or
  imprecise concepts,'' in \emph{Systems Theory in the Social Sciences}.\hskip
  1em plus 0.5em minus 0.4em\relax Springer, 1976, pp. 202--282.

\bibitem{zadeh1975conceptI}
------, ``The concept of a linguistic variable and its application to
  approximate reasoning—i,'' \emph{Information sciences}, vol.~8, no.~3, pp.
  199--249, 1975.

\bibitem{zadeh1975conceptII}
------, ``The concept of a linguistic variable and its application to
  approximate reasoning—ii,'' \emph{Information sciences}, vol.~8, no.~4, pp.
  301--357, 1975.

\bibitem{zadeh1975conceptIII}
------, ``The concept of a linguistic variable and its application to
  approximate reasoning-iii,'' \emph{Information sciences}, vol.~9, no.~1, pp.
  43--80, 1975.

\bibitem{zadeh1996fuzzy}
------, ``Fuzzy logic= computing with words,'' \emph{IEEE transactions on fuzzy
  systems}, vol.~4, no.~2, pp. 103--111, 1996.

\bibitem{zadeh1983computational}
------, ``A computational approach to fuzzy quantifiers in natural languages,''
  \emph{Computers \& Mathematics with applications}, vol.~9, no.~1, pp.
  149--184, 1983.

\bibitem{efstathiou1979multiattribute}
J.~Efstathiou and V.~Rajkovic, ``Multiattribute decisionmaking using a fuzzy
  heuristic approach,'' \emph{IEEE Transactions on Systems, Man, and
  Cybernetics}, vol.~9, no.~6, pp. 326--333, 1979.

\bibitem{chajda2015tense}
I.~Chajda and J.~Paseka, ``Tense operators in fuzzy logic,'' \emph{Fuzzy Sets
  and Systems}, vol. 276, pp. 100--113, 2015.

\bibitem{thiele1993fuzzy}
H.~Thiele and S.~Kalenka, ``On fuzzy temporal logic,'' in \emph{Fuzzy Systems,
  1993., Second IEEE International Conference on}.\hskip 1em plus 0.5em minus
  0.4em\relax IEEE, 1993, pp. 1027--1032.

\bibitem{moon2004fuzzy}
S.-i. Moon, K.~H. Lee, and D.~Lee, ``Fuzzy branching temporal logic,''
  \emph{IEEE Transactions on Systems, Man, and Cybernetics, Part B
  (Cybernetics)}, vol.~34, no.~2, pp. 1045--1055, 2004.

\bibitem{cardenas2006sound}
M.~Cardenas-Viedma, ``A sound and complete fuzzy temporal constraint logic,''
  \emph{IEEE Transactions on Systems, Man, and Cybernetics, Part B
  (Cybernetics)}, vol.~36, no.~1, pp. 223--228, 2006.

\bibitem{mukherjee2013fuzzy}
S.~Mukherjee and P.~Dasgupta, ``A fuzzy real-time temporal logic,''
  \emph{International Journal of Approximate Reasoning}, vol.~54, no.~9, pp.
  1452--1470, 2013.

\bibitem{prior2003time}
A.~N. Prior, \emph{Time and modality}.\hskip 1em plus 0.5em minus 0.4em\relax
  John Locke Lecture, 2003.

\bibitem{macfarlane2003future}
J.~MacFarlane, ``Future contingents and relative truth,'' \emph{The
  philosophical quarterly}, vol.~53, no. 212, pp. 321--336, 2003.

\bibitem{bellman1970decision}
R.~E. Bellman and L.~A. Zadeh, ``Decision-making in a fuzzy environment,''
  \emph{Management science}, vol.~17, no.~4, pp. B--141, 1970.

\bibitem{benzion1989discount}
U.~Benzion, A.~Rapoport, and J.~Yagil, ``Discount rates inferred from
  decisions: An experimental study,'' \emph{Management science}, vol.~35,
  no.~3, pp. 270--284, 1989.

\bibitem{chapman1996temporal}
G.~B. Chapman, ``Temporal discounting and utility for health and money.''
  \emph{Journal of Experimental Psychology: Learning, Memory, and Cognition},
  vol.~22, no.~3, p. 771, 1996.

\bibitem{chapman1995valuing}
G.~B. Chapman and A.~S. Elstein, ``Valuing the future: Temporal discounting of
  health and money,'' \emph{Medical decision making}, vol.~15, no.~4, pp.
  373--386, 1995.

\bibitem{pender1996discount}
J.~L. Pender, ``Discount rates and credit markets: Theory and evidence from
  rural india,'' \emph{Journal of development Economics}, vol.~50, no.~2, pp.
  257--296, 1996.

\bibitem{redelmeier1993time}
D.~A. Redelmeier and D.~N. Heller, ``Time preference in medical decision making
  and cost-effectiveness analysis,'' \emph{Medical Decision Making}, vol.~13,
  no.~3, pp. 212--217, 1993.

\bibitem{frederick2002time}
S.~Frederick, G.~Loewenstein, and T.~O'donoghue, ``Time discounting and time
  preference: A critical review,'' \emph{Journal of economic literature},
  vol.~40, no.~2, pp. 351--401, 2002.

\bibitem{read2001time}
D.~Read, ``Is time-discounting hyperbolic or subadditive?'' \emph{Journal of
  risk and uncertainty}, vol.~23, no.~1, pp. 5--32, 2001.

\bibitem{read2003subadditive}
D.~Read and P.~H. Roelofsma, ``Subadditive versus hyperbolic discounting: A
  comparison of choice and matching,'' \emph{Organizational Behavior and Human
  Decision Processes}, vol.~91, no.~2, pp. 140--153, 2003.

\bibitem{loewenstein1992anomalies}
G.~Loewenstein and D.~Prelec, ``Anomalies in intertemporal choice: Evidence and
  an interpretation,'' \emph{The Quarterly Journal of Economics}, vol. 107,
  no.~2, pp. 573--597, 1992.

\bibitem{ainslie2016cardinal}
G.~Ainslie, ``The cardinal anomalies that led to behavioral economics:
  Cognitive or motivational?'' \emph{Managerial and Decision Economics},
  vol.~37, no. 4-5, pp. 261--273, 2016.

\bibitem{loewenstein1989anomalies}
G.~Loewenstein and R.~H. Thaler, ``Anomalies: intertemporal choice,''
  \emph{Journal of Economic perspectives}, vol.~3, no.~4, pp. 181--193, 1989.

\bibitem{peters2016evaluating}
O.~Peters and M.~Gell-Mann, ``Evaluating gambles using dynamics,'' \emph{Chaos:
  An Interdisciplinary Journal of Nonlinear Science}, vol.~26, no.~2, p.
  023103, 2016.

\bibitem{peters2011time}
O.~Peters, ``The time resolution of the st petersburg paradox,'' \emph{Phil.
  Trans. R. Soc. A}, vol. 369, no. 1956, pp. 4913--4931, 2011.

\bibitem{bernoulli2011exposition}
D.~Bernoulli, ``Exposition of a new theory on the measurement of risk,'' in
  \emph{The Kelly Capital Growth Investment Criterion: Theory and
  Practice}.\hskip 1em plus 0.5em minus 0.4em\relax World Scientific, 2011, pp.
  11--24.

\bibitem{tversky1986rational}
A.~Tversky and D.~Kahneman, ``Rational choice and the framing of decisions,''
  \emph{Journal of business}, pp. S251--S278, 1986.

\bibitem{tversky1992advances}
------, ``Advances in prospect theory: Cumulative representation of
  uncertainty,'' \emph{Journal of Risk and uncertainty}, vol.~5, no.~4, pp.
  297--323, 1992.

\bibitem{kahneman2013prospect}
D.~Kahneman and A.~Tversky, ``Prospect theory: An analysis of decision under
  risk,'' in \emph{Handbook of the fundamentals of financial decision making:
  Part I}.\hskip 1em plus 0.5em minus 0.4em\relax World Scientific, 2013, pp.
  99--127.

\bibitem{kahneman2013choices}
------, ``Choices, values, and frames,'' in \emph{Handbook of the Fundamentals
  of Financial Decision Making: Part I}.\hskip 1em plus 0.5em minus 0.4em\relax
  World Scientific, 2013, pp. 269--278.

\bibitem{taleb2018skin}
N.~N. Taleb, \emph{Skin in the Game: Hidden Asymmetries in Daily Life}.\hskip
  1em plus 0.5em minus 0.4em\relax Random House, 2018.

\bibitem{lu2010many}
Z.~Lu, J.~Liu, J.~C. Augusto, and H.~Wang, ``A many-valued temporal logic and
  reasoning framework for decision making,'' in \emph{Computational
  Intelligence in Complex Decision Systems}.\hskip 1em plus 0.5em minus
  0.4em\relax Springer, 2010, pp. 125--146.

\bibitem{perry1965babrius}
B.~E. Perry \emph{et~al.}, ``Babrius and phaedrus,'' p. 487, 1965.

\bibitem{ohrstrom2007temporal}
P.~{\O}hrstr{\o}m and P.~Hasle, \emph{Temporal logic: from ancient ideas to
  artificial intelligence}.\hskip 1em plus 0.5em minus 0.4em\relax Springer
  Science \& Business Media, 2007, vol.~57.

\bibitem{mukherjee2017loss}
S.~Mukherjee, A.~Sahay, V.~C. Pammi, and N.~Srinivasan, ``Is loss-aversion
  magnitude-dependent? measuring prospective affective judgments regarding
  gains and losses,'' \emph{Judgment and Decision Making}, vol.~12, no.~1,
  p.~81, 2017.

\bibitem{kirby1997bidding}
K.~N. Kirby, ``Bidding on the future: Evidence against normative discounting of
  delayed rewards.'' \emph{Journal of Experimental Psychology: General}, vol.
  126, no.~1, p.~54, 1997.

\bibitem{kirby1995modeling}
K.~N. Kirby and N.~N. Marakovi{\'c}, ``Modeling myopic decisions: Evidence for
  hyperbolic delay-discounting within subjects and amounts,''
  \emph{Organizational Behavior and Human decision processes}, vol.~64, no.~1,
  pp. 22--30, 1995.

\bibitem{myerson1995discounting}
J.~Myerson and L.~Green, ``Discounting of delayed rewards: Models of individual
  choice,'' \emph{Journal of the experimental analysis of behavior}, vol.~64,
  no.~3, pp. 263--276, 1995.

\bibitem{rachlin1991subjective}
H.~Rachlin, A.~Raineri, and D.~Cross, ``Subjective probability and delay,''
  \emph{Journal of the experimental analysis of behavior}, vol.~55, no.~2, pp.
  233--244, 1991.

\bibitem{benhabib2004hyperbolic}
J.~Benhabib, A.~Bisin, and A.~Schotter, ``Hyperbolic discounting: An
  experimental analysis,'' in \emph{Society for Economic Dynamics Meeting
  Papers}, vol. 563, 2004.

\bibitem{laibson1997golden}
D.~Laibson, ``Golden eggs and hyperbolic discounting,'' \emph{The Quarterly
  Journal of Economics}, vol. 112, no.~2, pp. 443--478, 1997.

\bibitem{levy1997new}
M.~Levy and S.~Solomon, ``New evidence for the power-law distribution of
  wealth,'' \emph{Physica A: Statistical Mechanics and its Applications}, vol.
  242, no. 1-2, pp. 90--94, 1997.

\bibitem{druagulescu2001exponential}
A.~Dr{\u{a}}gulescu and V.~M. Yakovenko, ``Exponential and power-law
  probability distributions of wealth and income in the united kingdom and the
  united states,'' \emph{Physica A: Statistical Mechanics and its
  Applications}, vol. 299, no. 1-2, pp. 213--221, 2001.

\bibitem{sinha2006evidence}
S.~Sinha, ``Evidence for power-law tail of the wealth distribution in india,''
  \emph{Physica A: Statistical Mechanics and its Applications}, vol. 359, pp.
  555--562, 2006.

\bibitem{klass2006forbes}
O.~S. Klass, O.~Biham, M.~Levy, O.~Malcai, and S.~Solomon, ``The forbes 400 and
  the pareto wealth distribution,'' \emph{Economics Letters}, vol.~90, no.~2,
  pp. 290--295, 2006.

\bibitem{takahashi2006time}
T.~Takahashi, ``Time-estimation error following weber--fechner law may explain
  subadditive time-discounting,'' \emph{Medical hypotheses}, vol.~67, no.~6,
  pp. 1372--1374, 2006.

\bibitem{takahashi2008psychophysics}
T.~Takahashi, H.~Oono, and M.~H. Radford, ``Psychophysics of time perception
  and intertemporal choice models,'' \emph{Physica A: Statistical Mechanics and
  its Applications}, vol. 387, no. 8-9, pp. 2066--2074, 2008.

\bibitem{thaler1981some}
R.~Thaler, ``Some empirical evidence on dynamic inconsistency,''
  \emph{Economics letters}, vol.~8, no.~3, pp. 201--207, 1981.

\bibitem{ainslie1983motives}
G.~Ainslie and V.~Haendel, ``The motives of the will,'' \emph{Etiologic aspects
  of alcohol and drug abuse}, vol.~3, pp. 119--140, 1983.

\bibitem{green1994temporal}
L.~Green, N.~Fristoe, and J.~Myerson, ``Temporal discounting and preference
  reversals in choice between delayed outcomes,'' \emph{Psychonomic Bulletin \&
  Review}, vol.~1, no.~3, pp. 383--389, 1994.

\bibitem{holcomb1992another}
J.~H. Holcomb and P.~S. Nelson, ``Another experimental look at individual time
  preference,'' \emph{Rationality and Society}, vol.~4, no.~2, pp. 199--220,
  1992.

\bibitem{kirby1999heroin}
K.~N. Kirby, N.~M. Petry, and W.~K. Bickel, ``Heroin addicts have higher
  discount rates for delayed rewards than non-drug-using controls.''
  \emph{Journal of Experimental psychology: general}, vol. 128, no.~1, p.~78,
  1999.

\bibitem{loewenstein1987anticipation}
G.~Loewenstein, ``Anticipation and the valuation of delayed consumption,''
  \emph{The Economic Journal}, vol.~97, no. 387, pp. 666--684, 1987.

\bibitem{raineri1993effect}
A.~Raineri and H.~Rachlin, ``The effect of temporal constraints on the value of
  money and other commodities,'' \emph{Journal of Behavioral Decision Making},
  vol.~6, no.~2, pp. 77--94, 1993.

\bibitem{shelley1993outcome}
M.~K. Shelley, ``Outcome signs, question frames and discount rates,''
  \emph{Management Science}, vol.~39, no.~7, pp. 806--815, 1993.

\bibitem{green1997rate}
L.~Green, J.~Myerson, and E.~McFadden, ``Rate of temporal discounting decreases
  with amount of reward,'' \emph{Memory \& cognition}, vol.~25, no.~5, pp.
  715--723, 1997.

\bibitem{kirby1995preference}
K.~N. Kirby and R.~J. Herrnstein, ``Preference reversals due to myopic
  discounting of delayed reward,'' \emph{Psychological science}, vol.~6, no.~2,
  pp. 83--89, 1995.

\bibitem{millar1984self}
A.~Millar and D.~J. Navarick, ``Self-control and choice in humans: Effects of
  video game playing as a positive reinforcer,'' \emph{Learning and
  Motivation}, vol.~15, no.~2, pp. 203--218, 1984.

\bibitem{solnick1980experimental}
J.~V. Solnick, C.~H. Kannenberg, D.~A. Eckerman, and M.~B. Waller, ``An
  experimental analysis of impulsivity and impulse control in humans,''
  \emph{Learning and Motivation}, vol.~11, no.~1, pp. 61--77, 1980.

\bibitem{ainslie1981preference}
G.~Ainslie and R.~J. Herrnstein, ``Preference reversal and delayed
  reinforcement,'' \emph{Animal Learning \& Behavior}, vol.~9, no.~4, pp.
  476--482, 1981.

\bibitem{green1981preference}
L.~Green, E.~B. Fisher, S.~Perlow, and L.~Sherman, ``Preference reversal and
  self control: Choice as a function of reward amount and delay.''
  \emph{Behaviour Analysis Letters}, 1981.

\bibitem{nivanen2003generalized}
L.~Nivanen, A.~Le~Mehaute, and Q.~A. Wang, ``Generalized algebra within a
  nonextensive statistics,'' \emph{Reports on Mathematical Physics}, vol.~52,
  no.~3, pp. 437--444, 2003.

\bibitem{tsallis1988possible}
C.~Tsallis, ``Possible generalization of boltzmann-gibbs statistics,''
  \emph{Journal of statistical physics}, vol.~52, no. 1-2, pp. 479--487, 1988.

\bibitem{tsallis1999nonextensive}
------, ``Nonextensive statistics: theoretical, experimental and computational
  evidences and connections,'' \emph{Brazilian Journal of Physics}, vol.~29,
  no.~1, pp. 1--35, 1999.

\end{thebibliography}

\end{document}